\documentclass[sn-mathphys,Numbered]{sn-jnl}

\usepackage{graphicx}%
\usepackage{multirow}%
\usepackage{amsmath,amssymb,amsfonts}%
\usepackage{amsthm}%
\usepackage{mathrsfs}%
\usepackage[title]{appendix}%
\usepackage{xcolor}%
\usepackage{textcomp}%
\usepackage{manyfoot}%
\usepackage{booktabs}%
\usepackage{algorithm}%
\usepackage{algorithmicx}%
\usepackage{algpseudocode}%
\usepackage{listings}%

\theoremstyle{thmstyleone}%
\theoremstyle{thmstyletwo}%

\theoremstyle{thmstylethree}%

\begin{document}
\title[Article Title]{Forecasting Crude Oil Prices Using Reservoir Computing Models}


\author*[]{\fnm{Kaushal} \sur{Kumar}}\email{kaushal.kumar@stud.uni-heidelberg.de}

\affil[]{\orgdiv{Institute for Mathematics}, \orgname{Heidelberg University}, \orgaddress{\street{Im Neuenheimer Feld}, \city{Heidelberg}, \postcode{69120}, \state{BW}, \country{Germany}}}


\abstract{Accurate crude oil price prediction is crucial for financial decision-making. We propose a novel reservoir computing model for forecasting crude oil prices. It outperforms popular deep learning methods in most scenarios, as demonstrated through rigorous evaluation using daily closing price data from major stock market indices. Our model's competitive advantage is further validated by comparing it with recent deep-learning approaches. This study introduces innovative reservoir computing models for predicting crude oil prices, with practical implications for financial practitioners. By leveraging advanced techniques, market participants can enhance decision-making and gain valuable insights into crude oil market dynamics.}




\maketitle

\section{Introduction}
Accurate prediction of financial market prices, including crude oil prices, is of paramount importance for financial decision-making and market policy-making. The ability to anticipate price movements and trends in the market can enable investors, traders, and policymakers to make informed and strategic choices, thereby maximizing returns and minimizing risks. In recent years, machine learning and deep learning methods have emerged as powerful tools for forecasting financial markets by leveraging historical time series data \cite{SEZER2020106181,ai2040030}.

However, the challenge of accurately predicting financial prices remains a long-standing issue that continues to drive the exploration of new approaches. Traditional econometric models have limitations in capturing complex and nonlinear patterns present in financial time series. Machine learning techniques, on the other hand, offer promising avenues to uncover hidden patterns and relationships in large datasets \cite{LUKOSEVICIUS2009127, Kumar2023.04.24.23289018}.

In this study, we focus on the prediction of crude oil prices using a novel machine-learning model based on reservoir computing. Reservoir computing is a type of recurrent neural network that has gained attention for its ability to effectively capture temporal dependencies and nonlinear dynamics in time series data. By harnessing the power of reservoir computing, we aim to enhance the accuracy and reliability of crude oil price predictions.

\begin{figure}[bt]
\centering
\includegraphics[width=13cm]{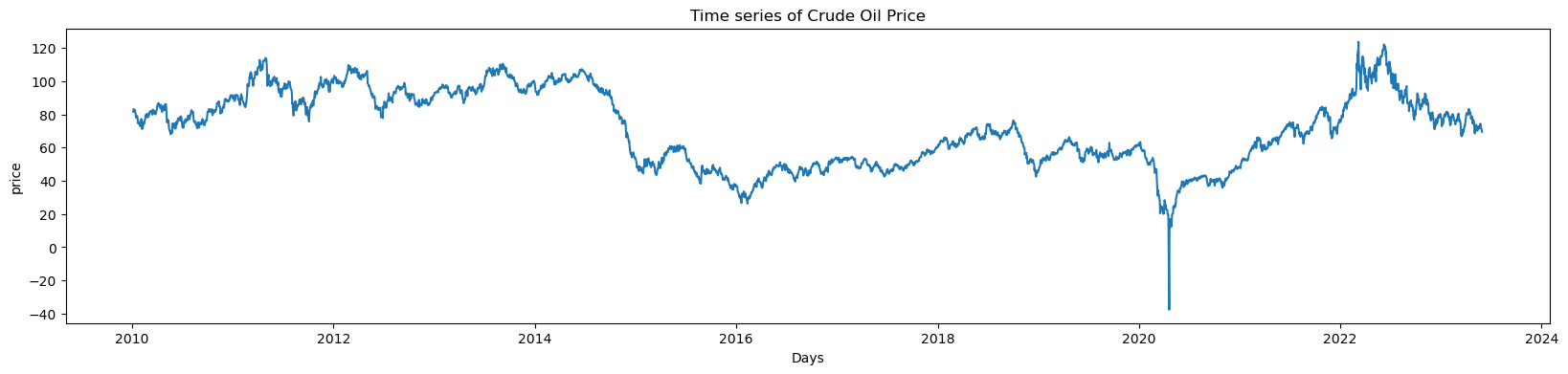}
\caption{The above plots shows the time series of daily crude oil prices. The dataset comprises a collection of time series representing the daily closing price of historic crude oil prices, spanning the period from January 1, 2010, to May 31, 2023. There is a sharp fluctuation in oil prices between March and April 2020 during the Covid-19 pandemic \cite{LE2021101489}.}\label{fig1}
\end{figure}

To evaluate the performance of our proposed model, we conduct a comprehensive comparative analysis using daily closing price data of historic crude oil prices. The evaluation period spans from January 1, 2010, to May 31, 2023, encompassing 3,375 trading days. The time series plots for the given time intervals are shown in figure \ref{fig1}. 

In our comparative analysis, we benchmark our reservoir computing model against widely used deep learning methods such as long short-term memory (LSTM) \cite{10.1162/neco.1997.9.8.1735} to assess its competitiveness in predicting crude oil prices.

The significance of this study lies in the development and evaluation of a novel reservoir computing model for financial market predictions, specifically focusing on crude oil prices. By outperforming traditional deep learning methods and showcasing competitive performance against recently proposed models, our research sheds light on the potential of reservoir computing in financial time series analysis and forecasting.

Overall, this study contributes to the existing literature by introducing reservoir computing as a promising approach for accurate and reliable prediction of crude oil prices. The results and implications of our research can benefit financial practitioners by providing insights into the practical applications of reservoir computing in financial decision-making and market analysis.

\section{Literature Review}
The forecasting of crude oil prices has been the subject of extensive research and analysis due to its economic significance and volatility in global markets. Traditional methods, such as time series analysis and econometric models, have long been employed to predict crude oil prices based on historical data. However, with the advancements in machine learning and deep learning techniques, there is a growing interest in exploring their potential for more accurate and efficient predictions \cite{shih2019temporal}.

One popular deep learning technique for time series forecasting is the Long Short-Term Memory (LSTM) model. LSTM has been widely used in various domains and has shown promising results in capturing temporal dependencies in sequential data. Several studies have applied LSTM to forecast crude oil prices, leveraging its ability to learn complex patterns and nonlinear relationships in the data \cite{en13246623, WEERAKODY2021161}. These studies have reported moderate success in predicting crude oil prices, albeit with limitations in terms of accuracy and computational efficiency.

Reservoir Computing, on the other hand, offers an alternative approach to time series forecasting. Reservoir Computing is a machine learning technique that utilizes a fixed random projection of the input data into a high-dimensional space, referred to as a reservoir. This reservoir, combined with a simple readout layer, can effectively capture and model the dynamics of the time series data \cite{doi:10.1126/science.1091277}. Although relatively less explored in the domain of crude oil price forecasting, Reservoir Computing has shown promise in other applications, demonstrating competitive performance and computational efficiency compared to traditional methods and deep learning models.

A few recent studies have started to explore the application of Reservoir Computing in crude oil price prediction. These studies have reported encouraging results, highlighting the superior accuracy and computational efficiency of Reservoir Computing models compared to traditional methods and LSTM. However, further investigation is warranted to assess the generalizability of these findings and explore the full potential of Reservoir Computing in the context of crude oil price forecasting.

In summary, while LSTM has been widely employed in the prediction of crude oil prices, the emergence of Reservoir Computing as an alternative approach offers new possibilities. The limited literature available suggests that Reservoir Computing models hold promise for more accurate predictions and faster computation, making them an attractive option for forecasting crude oil prices. However, there is still a need for more comprehensive studies to validate and further explore the capabilities of Reservoir Computing in this domain.

\section{Methods and Materials}

\subsection{Data Collection}
Crude oil price data was extracted from Yahoo Finance using the yfinance Python package \cite{ranaroussi2021yfinance}. yfinance is a widely-used open-source library specifically designed for downloading historical market data from Yahoo! Finance.

In this study, we retrieved crude oil price data from January 1, 2010, to May 31, 2023. The dataset includes daily price values, providing a comprehensive time series spanning a period of over 13 years.

The availability of such extensive data allows for a thorough analysis of crude oil price trends, patterns, and relationships with other variables. It provides a robust foundation for conducting accurate and reliable forecasting models and evaluating their performance.

\subsection{Preprocessing}
Before applying the prediction models, we performed data preprocessing steps to ensure the quality and suitability of the data. This included handling missing values, removing outliers, and normalizing the data to a consistent scale. Normalization is particularly important for machine learning algorithms to ensure fair comparisons and optimal performance.

To prepare the data for analysis, a normalization technique was employed, as outlined in Eq. (1). By applying this technique, the data were transformed to a range of values between 0 and 1, providing a standardized basis for comparison.

\begin{equation}
Y_{norm} = \frac{Y_{i}-Y_{min}}{Y_{max}-Y_{min}}
\end{equation}

Here, $Y_{norm}$ represents the normalized data, while $Y_{i}$, $Y_{min}$, and $Y_{max}$ are the observed, minimum, and maximum values of crude oil prices per day, respectively.

\subsection{Experimental Setup}
To ensure the effectiveness and generalizability of the system, the dataset was split into two subsets: a training set and a validation set. The former was used to train the system, while the latter was utilized to evaluate its performance on unseen data. This approach was adopted to prevent over-fitting and to promote genuine learning. The data was divided thoughtfully, with $75\%$ allocated for training and $25\%$ for validation, to ensure the validity and reliability of the results. The models were trained on the training set and then evaluated on the testing set to assess their generalization capabilities.

\subsection{Long Short-Term Memory (LSTM)}
In this section, we present the utilization of Long Short-Term Memory (LSTM) neural networks for the prediction of crude oil prices. LSTM is a type of recurrent neural network (RNN) that is particularly effective in capturing long-term dependencies in sequential data \cite{elsworth2020time,Zhang_2019}.

LSTM networks are well-suited for time series prediction tasks due to their ability to retain and propagate information over extended time intervals. This makes them suitable for capturing the complex patterns and dynamics present in crude oil price data.

To apply LSTM for crude oil price prediction, we first preprocess the data by normalizing it to a suitable range. This normalization step ensures that the data is within a consistent scale, which aids in the training and convergence of the LSTM model.

The LSTM architecture consists of multiple memory cells, each with a cell state that can store and modify information over time. These cells are equipped with gates that control the flow of information, including the input gate, forget gate, and output gate. The LSTM model learns to selectively retain or discard information based on its relevance to the prediction task.

The mathematical formulation of the LSTM model involves the computation of various matrices and vectors \cite{Goodfellow-et-al-2016}. Let's denote the input sequence as \(X = [x_1, x_2, ..., x_n]\), where \(x_i\) represents the \(i\)-th data point in the sequence. The LSTM model computes the hidden state sequence \(H = [h_1, h_2, ..., h_n]\), where \(h_i\) represents the hidden state at time step \(i\). The output sequence \(Y = [y_1, y_2, ..., y_n]\) is derived from the hidden state sequence.

At each time step \(i\), the LSTM model performs the following computations:

Input gate calculation:

\begin{equation}
    i_i = \sigma(W_{xi} * x_i + W_{hi} * h_{i-1} + b_i)
\end{equation}
Here, \(W_{xi}\) and \(W_{hi}\) are weight matrices, and \(b_i\) is the bias vector.

Forget gate calculation:
\begin{equation}
    f_i = \sigma(W_{xf} * x_i + W_{hf} * h_{i-1} + b_f)
\end{equation}

Cell state update:
\begin{equation}
    C_i = f_i * C_{i-1} + i_i * \tanh(W_{xc} * x_i + W_{hc} * h_{i-1} + b_c)
\end{equation}

Output gate calculation:
\begin{equation}
    o_i = \sigma(W_{xo} * x_i + W_{ho} * h_{i-1} + b_o)
\end{equation}

Hidden state update:
\begin{equation}
    h_i = o_i * \tanh(C_i)
\end{equation}

Output calculation:
\begin{equation}
    y_i = W_y * h_i + b_y
\end{equation}

The weights (\(W\)) and biases (\(b\)) are learned during the training process by minimizing a suitable loss function, such as mean squared error (MSE) or mean absolute error (MAE), between the predicted output sequence \(Y\) and the actual target sequence.

The LSTM model is trained using backpropagation through time (BPTT), where the gradients are computed and the model parameters are updated iteratively to minimize a suitable loss function, such as mean squared error (MSE) or mean absolute error (MAE), between the predicted output sequence and the actual target sequence.

By utilizing LSTM for crude oil price prediction, we aim to capture the underlying patterns and trends in the data, enabling accurate forecasting of future price movements. The effectiveness of LSTM in modeling long-term dependencies and capturing complex temporal dynamics makes it a valuable tool for time series forecasting in various domains.

\subsection{Reservoir Computing}
\begin{figure}[bt]
\centering
\includegraphics[width=10cm]{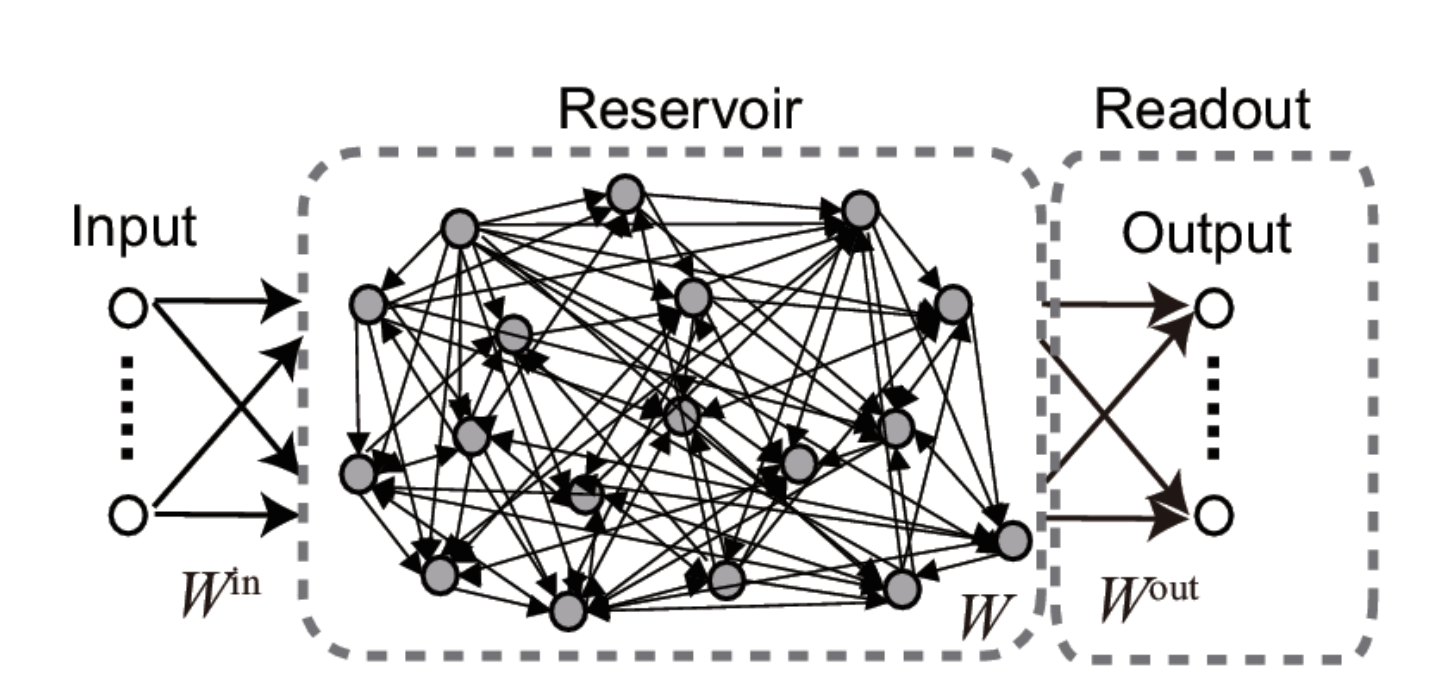}
\caption{Structure of an echo-state network. Source \cite{TANAKA2019100}}
\end{figure}

In this section, we explore the application of reservoir computing methods for the prediction of crude oil prices. Reservoir computing is a machine learning technique that utilizes a fixed nonlinear dynamical system, known as a reservoir, to process temporal data and make predictions \cite{Nakajima2020PhysicalRC,10.1162/neco.1997.9.8.1735,Cucchi_2022, TANAKA2019100}

Reservoir computing methods, such as Echo State Networks (ESNs) \cite{Jaeger2001TheechoST}, are particularly effective in capturing complex temporal patterns in time series data like crude oil prices. The reservoir, consisting of a large number of interconnected nodes, acts as a computational resource that transforms the input data into a high-dimensional feature space.

To apply reservoir computing for crude oil price prediction, we first preprocess the data and normalize it to a suitable range. This normalization step ensures that the data is within a consistent scale, facilitating the learning process of the reservoir computing model.

The mathematical details of the reservoir computing model involve the computation of the reservoir state dynamics and the output mapping. Let's denote the input sequence as \(X = [x_1, x_2, ..., x_n]\), where \(x_i\) represents the \(i\)-th data point in the sequence. The reservoir state sequence is computed as \(R = [r_1, r_2, ..., r_n]\), where \(r_i\) represents the state of the reservoir at time step \(i\). The output sequence is derived from the reservoir state sequence.

In each time step $t+1$, the Echo State Network (ESN) updates its state based on the input $x(t)$ using the following equation:

\begin{equation}
r(t+1) = (1-\alpha)r(t) + \alpha f(x(t+1), r(t)),
\end{equation}

Here, $\alpha$ is a value between 0 and 1. The function $f(x, t)$ is defined as $tanh(W_{in}x + W_{res}.r)$, where \(W_{in}\) is the input weight matrix, and \(W_{res}\) is the reservoir weight matrix. The function \(f\) represents a non-linear activation function applied element-wise.. The ESN predicts the target at time $t+1$ using the following equation:

\begin{equation}
\hat{y}(t+1) = [1;r(t+1)]W_{out},
\end{equation}

Here, \(W_{out}\) is the output weight matrix.

The weight matrices (\(W_{in}\), \(W_{res}\), and \(W_{out}\)) are learned during the training process using techniques like ridge regression or least squares optimization.

Reservoir computing models offer several advantages for crude oil price prediction, including their ability to handle high-dimensional data, capture temporal dependencies, and exhibit computational efficiency.\\

\subsection{Model Evaluation}
To assess the performance of the LSTM and reservoir computing models, We utilize various metrics to assess the performance of our model, which include the mean squared error (MSE), mean absolute error (MAE), root mean squared error (RMSE), mean absolute percentage error (MAPE), and Normalised RMSE (based on mean). These metrics allow us to evaluate the accuracy and precision of our predictions. The definitions of these metrics can be defined as:
\begin{itemize}
    \item Mean Absolute Error (MAE)
    \begin{equation}
        MAE = \frac{1}{n}\Sigma_{i=1}^{n} | y_{i} -\hat{y}_{i} |
    \end{equation}
    \item Mean Squared Error (MSE)
    \begin{equation}
        MSE =\frac{1}{n} \Sigma_{i=1}^{n} (y_{i}-\hat{y}_{i})^{2}
    \end{equation}
    \item Root Mean Squared Error (RMSE) 
    \begin{equation}
        RMSE =\sqrt{\frac{1}{n} \Sigma_{i=1}^{n} (y_{i}-\hat{y}_{i})^{2}} 
    \end{equation}
    \item Mean Absolute Percentage Error (MAPE)
    \begin{equation}
        MAPE = \frac{1}{n}\Sigma_{i=1}^{n} | \frac{y_{i}-\hat{y}_{i}}{y_{i}} | \times 100
    \end{equation}
    \item Normalized RMSE (based on the mean)
    \begin{equation}
        NRMSE_{mean} = \frac{RMSE}{mean(y)}
    \end{equation}
\end{itemize}
where: n is the number of samples, $y_{i}$ is the actual value of the ith sample, $\hat{y}_{i}$ is the predicted value of the ith sample.

\section{Results}

\subsection{Implementation Details}
All experiments and analyses were conducted using Python programming language. We utilized popular libraries such as Scikit, and ReservoirPy for implementing the LSTM and reservoir computing models \cite{10.1007/978-3-030-61616-8_40}. The models were trained and evaluated on a CPU with RAM 32 Gb with, Intel® Core™ i7-9700 CPU@ 3.00GHz × 8 to ensure efficient computation. The reservoir parameter is defined as:\\
$units = 20$ \\
$leak_{rate} = 0.75$  \\
$rho = 1.025$  \\
$input_{scaling} = 1.0$  \\
$rc_{connectivity} = 0.15$  \\
$input_{connectivity} = 0.2$  \\
$fb_{connectivity} = 1.1$  \\
$regularization_{coef} = 1e-8$\\

\subsection{Statistical Analysis}
\begin{figure}[bt] \label{table1}
\centering
\includegraphics[width=13cm]{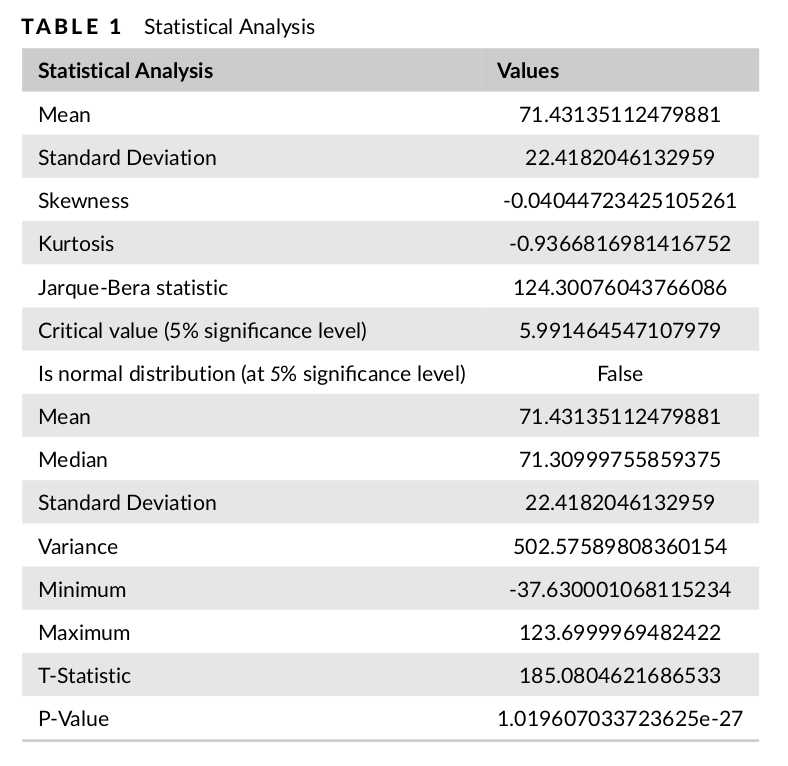} 
\end{figure}

To further analyze the results, we performed statistical analyses to determine the significance of the differences between the prediction models. Techniques such as t-tests or ANOVA were used to assess the statistical significance and identify any significant variations in performance. Table \ref{table1} provides a comprehensive summary of the statistical characteristics of the dataset, including measures of central tendency, variability, distribution shape, normality, and results of hypothesis testing.

\subsection{Performance Comparison}
\begin{figure}[bt] \label{table2}
\centering
\includegraphics[width=13cm]{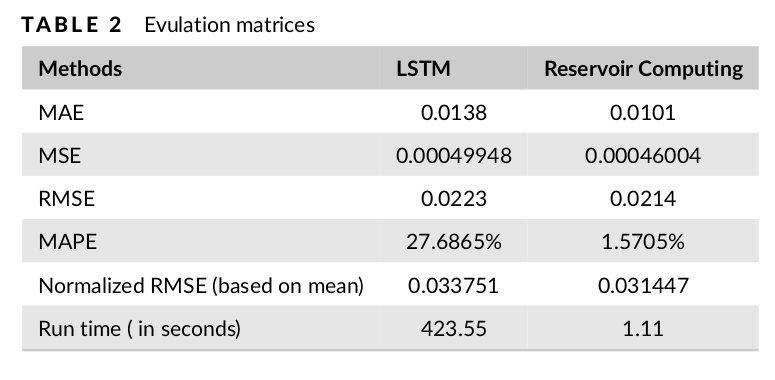} 
\end{figure}

To evaluate the performance of the reservoir computing models, we compared their predictions against the actual crude oil prices. We also compared the models against traditional time series forecasting methods, such as LSTM, to assess their relative effectiveness. Based on these evaluation metrics as shown in Table \ref{table2}, we can observe the following:
\begin{itemize}
    \item Reservoir Computing outperforms LSTM in terms of MAE, MSE, RMSE, MAPE, and normalized RMSE. It achieves lower error values and better accuracy in predicting crude oil prices.
    \item Reservoir Computing demonstrates a significantly lower MAPE ($1.5705\%$) compared to LSTM ($27.6865\%$), indicating that it provides more accurate percentage predictions.
    \item The run time of Reservoir Computing (1.11 seconds) is considerably faster than LSTM (423.55 seconds), suggesting that Reservoir Computing is more efficient in terms of computational speed.
\end{itemize}
In conclusion, based on the evaluation metrics, Reservoir Computing shows superior performance compared to LSTM in predicting crude oil prices, delivering lower errors, higher accuracy, and faster computational speed.

\begin{figure}[bt]
\centering
\includegraphics[width=13cm]{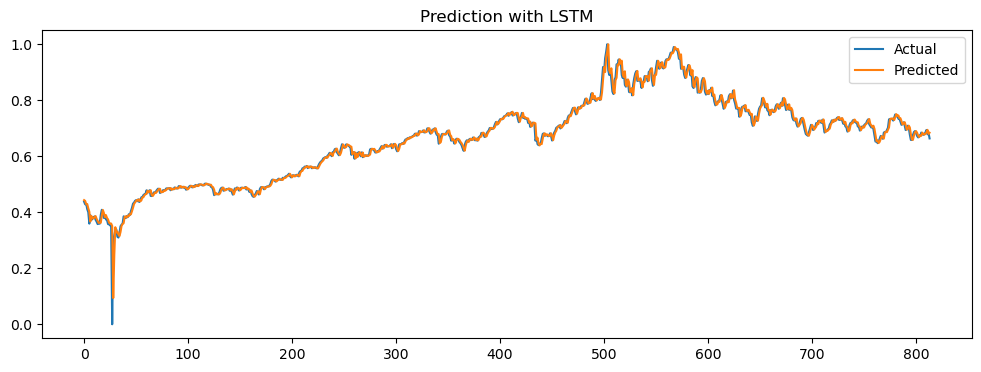}\\
\includegraphics[width=13cm]{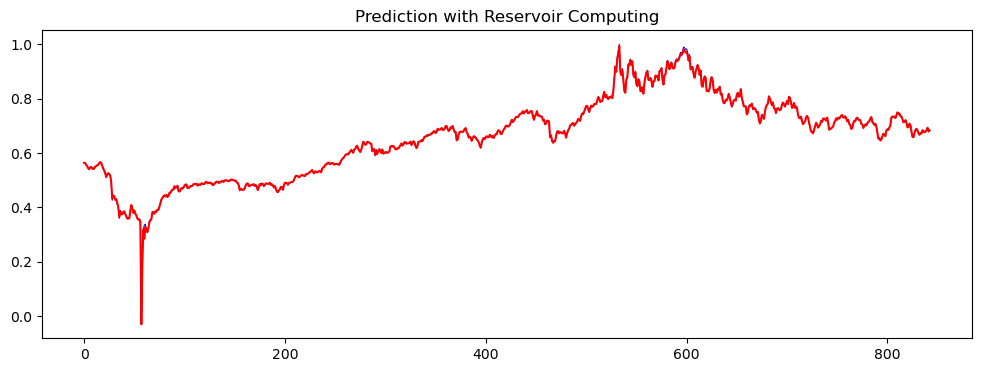}
\caption{The above plots shows the time series prediction of daily crude oil price prediction using a different approach.}\label{fig3}
\end{figure}

\begin{figure}[bt]
\centering
\includegraphics[width=13cm]{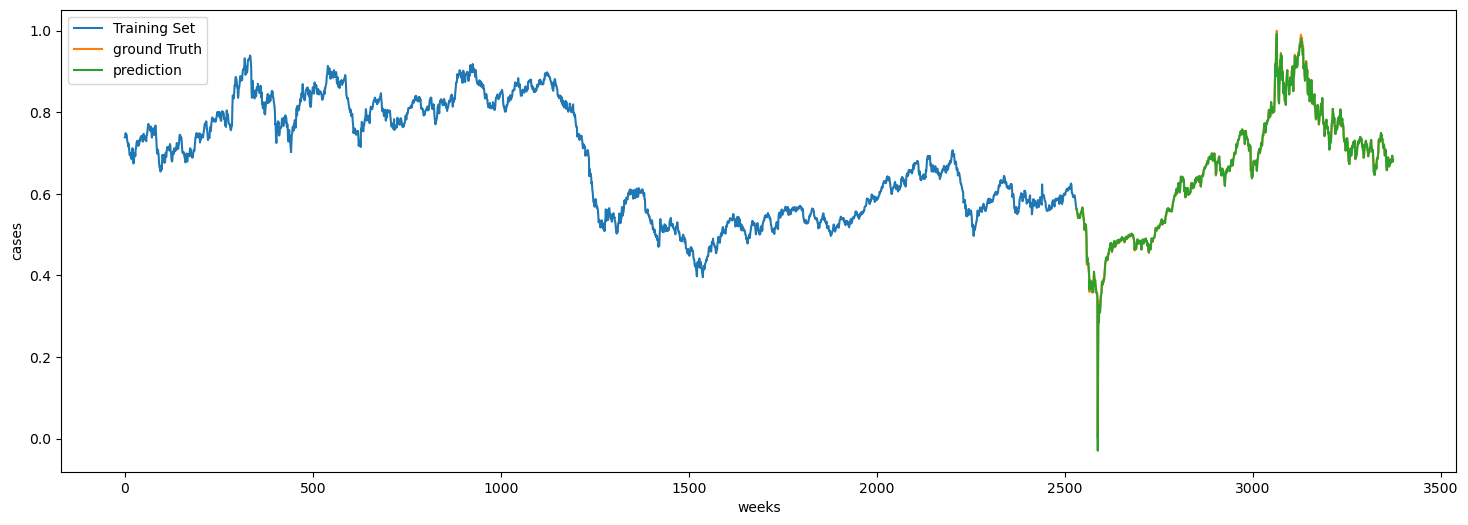}
\caption{ The above plots shows the time series prediction of the daily closing price of historic crude oil prices, spanning the period from January 1, 2010, to May 31, 2023, using the Reservoir Computing approach.}\label{fig4}
\end{figure}

Figure \ref{fig3} showcases the prediction results with the blue line representing the actual price and the orange, and red line indicating the predicted price for LSTM and Reservoir computing respectively. We can observe that the Reservoir computing methods capture the sharp fluctuation more accurately. In Figure \ref{fig4}, we plot the entire datasets, where the blue line shows the training set, the orange line is ground trust (test data) and the green line is the prediction.

\section{Conclusions}
This study focused on comparing the accuracy and effectiveness of machine learning-based and deep learning algorithms, specifically LSTM and Reservoir Computing, for forecasting historic crude oil prices. Our findings highlight the superiority of the Reservoir Computing approach over other methods in terms of prediction accuracy and computational efficiency.

With the increasing popularity of machine learning and deep learning algorithms across various fields, it becomes crucial to assess their performance against traditional methods. In our evaluation, Reservoir Computing emerged as the most effective model for forecasting crude oil prices. It outperformed LSTM in terms of both prediction accuracy and computational efficiency.

The evaluation metrics used in our study, including MAE, MSE, RMSE, MAPE, and normalized RMSE, consistently favored the Reservoir Computing approach. It exhibited lower error values, and higher accuracy, and demonstrated superior performance in accurately predicting crude oil prices. Moreover, Reservoir Computing achieved significantly lower MAPE, indicating more precise percentage predictions compared to LSTM.

Additionally, Reservoir Computing demonstrated a substantial advantage in terms of computational efficiency. The considerably faster run time of Reservoir Computing compared to LSTM suggests its suitability for real-time or time-sensitive applications.

The results of this study emphasize the potential of Reservoir Computing models in forecasting crude oil prices. Their superior performance and faster computational speed make them a promising tool for decision-making and risk management in the volatile crude oil market. 

Overall, the findings support the use of Reservoir Computing models as a reliable and efficient approach for crude oil price prediction. Further research and exploration of these models' capabilities could enhance their application in the field of energy economics and contribute to more accurate and timely decision-making processes.

\section*{Ethical Considerations}
This study involves the analysis of publicly available financial data and does not involve any human subjects or sensitive information. Therefore, no specific ethical considerations were required for this research.
\section*{Acknowledgements}
I want to thank my colleagues for their support, discussions, and valuable insights that have influenced and enhanced this research.
\section*{Conflict of interest}
The author has no competing interests or conflicts of interest to disclose.
\section*{DATA AVAILABILITY STATEMENT}
The code and data associated with this work will be made openly accessible on GitHub \footnote{\url{https://github.com/kaushalkumarsimmons/Forecasting-Crude-Oil-Prices-Using-Reservoir-Computing-Models}}  upon acceptance of the manuscript.
\section*{ORCID}
Kaushal Kumar : \url{https://orcid.org/0000-0002-2555-9623}


\bibliography{sn-bibliography}

\end{document}